\documentclass[final,5p,times,twocolumn]{elsarticle}
\usepackage{macros}
\begin{document}
%%%%%%%%%%%%%%%%%%%%%%%%%%%%%%%%%%%%%%%%%%%%%%%%%%%%%%%%%%%%%%%%%%
\begin{frontmatter}
%%%%%%%%%%%%%%%%%%%%%%%%%%%%%%%%%%%%%%%%%%%%%%%%%%%%%%%%%%%%%%%%%%
\title{Accelerating Knowledge Graph and Ontology Engineering with Large Language Models}
%%%%%%%%%%%%%%%%%%%%%%%%%%%%%%%%%%%%%%%%%%%%%%%%%%%%%%%%%%%%%%%%%%
\author[wsu]{Cogan Shimizu}
\ead{cogan.shimizu@wright.edu}
\author[ksu]{Pascal Hitzler}
\ead{hitzler@ksu.edu}
%%%%%%%%%%%%%%%%%%%%%%%%%%%%%%%%%%%%%%%%%%%%%%%%%%%%%%%%%%%%%%%%%%
\address[wsu]{Department of Computer Science and Engineering, Wright State University, USA}
\address[ksu]{Department of Computer Science, Kansas State University, USA}
%%%%%%%%%%%%%%%%%%%%%%%%%%%%%%%%%%%%%%%%%%%%%%%%%%%%%%%%%%%%%%%%%%
\begin{abstract}
    Large Language Models bear the promise of significant acceleration of key Knowledge Graph and Ontology Engineering tasks, including ontology modeling, extension, modification, population, alignment, as well as entity disambiguation. We lay out LLM-based Knowledge Graph and Ontology Engineering as a new and coming area of research, and argue that modular approaches to ontologies will be of central importance.
\end{abstract}
%%%%%%%%%%%%%%%%%%%%%%%%%%%%%%%%%%%%%%%%%%%%%%%%%%%%%%%%%%%%%%%%%%
%%%%%%%%%%%%%%%%%%%%%%%%%%%%%%%%%%%%%%%%%%%%%%%%%%%%%%%%%%%%%%%%%%
\end{frontmatter}
%%%%%%%%%%%%%%%%%%%%%%%%%%%%%%%%%%%%%%%%%%%%%%%%%%%%%%%%%%%%%%%%%%
\section{Introduction}
\label{sec:intro}
%%%%%
Knowledge Graph and Ontology Engineering (KGOE, in short) refers to a (vaguely defined) set of tasks that are of central relevance to the life cycle of knowledge graphs, and of ontologies,\footnote{We mostly understand ontologies as a type of \emph{schema} for knowledge graphs in the sense of \cite{momo}, and as such the ontology can in fact also be understood to be \emph{part} of the knowledge graph. We acknowledge other uses, to which our discussion also applies.} as data artifacts used in data management and applications.\footnote{For background on this and the more general Semantic Web field, see, e.g., \cite{hitzler-cacm} and the references given therein.} These include, for example, ontology modeling (i.e., construction), ontology population (i.e., creating a knowledge graph with the given ontology as schema), ontology extension and modification, ontology alignment, entity disambiguation (sometimes called co-reference resolution). 

All of the just mentioned tasks have in common that they are \emph{hard}, in the sense that even after more than a quarter century of Semantic Web research, they still defy attempts to automate them at reasonable quality levels and scale. The state of the art on all of these is that they require significant human expert labor, at times (such as for entity disambiguation) together with detailed scripting of algorithms that solve the problem at scale but only for a specific problem instance, i.e., for a very specific knowledge graph and/or ontology.

At the same time, knowledge graphs and ontologies are ever more important for applications in data integration and data management, and more recently also as ground truth to escape from Large Language Model (LLM) hallucinations \cite{lewis2020retrieval} and as components of other neurosymbolic approaches \cite{NeSyKG-book,DBLP:series/faia/342,DBLP:series/faia/369}. As a consequence, improved processes and methods for automating or even semi-automating core KGOE tasks remains a key challenge for the research community.

LLMs enter the scene, and the public perception of Artificial Intelligence (AI), with force in 2022 at the launch of OpenAI's ChatGPT,\footnote{\url{https://openai.com/index/hello-gpt-4o/}} with rapid developments since then. Their human-style conversation capabilities, which include a profound mastery of expression in written language, as well as solid production of more structured information, are as stunning as their sometimes wildly confabulated responses (usually termed \emph{hallucinations}). Perhaps most important for our discussion herein, LLMs appear to capture, and to have the ability to recall in different formats and contexts, a wide swath of human knowledge, both commonsense and specialized, provided it is reflected well enough in the training data. While at this point in time, their reliability in terms of accuracy of content in their responses remains problematic, it is quite apparent and widely reported that working with an LLM can save significant time and effort provided there is a (human) topic expert available as a check on factual accuracy.

The promise of LLMs for KGOE is thus: by using LLMs as approximate natural language knowledge bases that can be approximately queried, plus LLM capabilities to understand and produce information in ways structured for KGOE use, it should be possible to design semi-automatic methods, or human-LLM interactive methods, that can produce at least \emph{draft solutions} for key KGOE problems at a level of quality that will significantly reduce human expert time and effort. Work on this line of research has of course already been started by the Semantic Web community. With this paper, we intend to begin to consolidate the discussion, and -- in particular -- contribute observations and discussions related to our own research on modular ontologies, which we believe to be highly relevant, for reasons that we will lay out below.

The plan of this paper is as follows. In Section \ref{sec:divide} we motivate the need for (a certain type of) modularity for LLM-based KGOE. In Section \ref{sec:added} we discuss our notion of modularity and its research context. In Section \ref{sec:modular} we briefly look, in turn, at the key KGOE tasks identified in the introduction. In Section \ref{sec:challenges} we will list some concrete research challenges that can drive LLM-based KGOE forward, and in Section \ref{sec:conc} we will conclude.

%%%%%%%%%%%%%%%%%%%%%%%%%%%%%%%%%%%%%%%%%%%%%%%%%%%%%%%%%%%%%%%%%%
\section{Divide and Conquer}
\label{sec:divide}
%%%%%
Useful ontologies and knowledge graphs often tend to be large, or even very large. The ENVO ontology \cite{DBLP:journals/biomedsem/ButtigiegPLSWM16} has over $2,100$ classes; the Gene Ontology \cite{DBLP:journals/nar/CarbonDGUHMBCDH21} has over $40,000$ classes; a Wikipedia-derived class hierarchy used for explainable deep learning ana\-ly\-sis has over $2 \cdot 10^6$ classes \cite{DBLP:conf/kgswc/SarkerSHZNMJRA20}; DBpedia Core\footnote{\url{https://www.dbpedia.org/resources/latest-core/}} has about $9\cdot 10^8$ triples with an ontology of about 800 classes, KnowWhereGraph \cite{shimizu2024knowwheregraphontology} has over $28\cdot 10^9$ triples with an ontology of about 300 classes. The size -- in particular of the ontology underlying a knowledge graph -- is one of the obstacles facing humans in KGOE tasks, as even an ontology with a ``mere'' 300 classes (and corresponding comprehensive OWL axiomatization) is simply too big for a human to keep an overview of as a whole, or to understand well from scratch within a reasonable amount of time. 

Even before the advent of LLMs, therefore, Semantic Web researchers have looked into the use of ontology modules in order to segment large ontologies into pieces of a size that are more workable for the human brain. These efforts include this paper's authors and their collaborators, in a line of work that has developed out of Ontology Design Patterns (ODPs \cite{eva-dissertation,Gangemi05}) research and practice, using an approach that we believe to be particularly suited for LLM-based KGOE, and that we will focus on further below: If an ontology is split into (even possibly overlapping) pieces that make conceptual sense (as a type of mini-ontology) for the domain experts, then KGOE tasks can often focus on one or a few coherent pieces (modules) of the ontology, thus simplifying the task significantly. And given the frequent difficulties (and expense) of LLMs to deal with substantial size prompts and with more open-ended and less common scenarios, it is a reasonable expectation that similar benefits will apply to LLM-based KGOE.

Let us provide a case in point from the context of ontology alignment, reported in \cite{KGSWC-OA-2024}. The setting is the creation of complex ontology mapping rules between two ontologies in the OAEI GeoLink complex ontology alignment benchmark \cite{cob-geo}. The benchmark is of moderate size, with 40 classes, 149 object properties and 49 data properties in one of the ontology, and 156 classes, 124 object properties and 46 data properties in the other ontology. It is a natural rather than synthetic benchmark in that the ontologies were originally developed for an application purpose, and were only cast into a benchmark later. The benchmark (like all complex ontology alignment benchmarks) had defied automation for years, the only approaches that were able to create reasonably good results assumed a shared ABox (i.e. a shared data graph) \cite{DBLP:conf/semweb/PourAAFFHHJJKKL20,DBLP:conf/semweb/PourAAAFFFHHHHI21}, which is of course a very unrealistic assumption for data management practice.

As reported in \cite{KGSWC-OA-2024}, an LLM was prompted to produce a body of an alignment rule, given a rule head together with the two ontologies as part of the prompt. This essentially completely failed, i.e., the LLM produced output that was essentially unusable. But then we made use of the modular structure of the body-side ontology, for which 20 named modules (such as ``Organization'' or ``Physical Sample'') had already been provided at original deployment of the ontology \cite{geolink-ont}. We thus prompted in two stages: given an alignment rule head and the list of 20 module names, we first asked for the module(s) that would be required to create the rule. Then we prompted for the rule body by providing the modules previously identified. The results we obtained were really good in terms of high precision and recall. The availability and principled use of modules made the difference between an almost complete failure and a reasonably accurate system response. 

%%%%%%%%%%%%%%%%%%%%%%%%%%%%%%%%%%%%%%%%%%%%%%%%%%%%%%%%%%%%%%%%%%
\section{Modules}
\label{sec:added}
%%%%%
The term \emph{module} has been used for different things in the Semantic Web context.\footnote{See, e.g., the proceedings of the WoMO -- Workshops on Modular Ontologies, details at \url{https://iaoa.org/womo/history.html}.} Our approach, as in \cite{momo}, comes out of the tradition of ODPs \cite{eva-dissertation,Gangemi05}, which are partial ontologies that address commonly occurring, domain invariant ontology design issues. Later, related notions with similar underlying ideas, but generally different objectives (and implementations), include \cite{dosp,ottr}. 

For us, and this is laid out in \cite{modont,momo}, a module is a part of an ontology that consists of the classes, properties, and axioms within that ontology that are relevant to a key notion, as considered by a domain expert, for the ontology use case. For example, the above mentioned GeoLink ontologies that focus on oceanographic cruises and data, such key notions (and corresponding modules) include \emph{Trajectory}, \emph{Cruise}, \emph{Physical Sample}, \emph{Organization}. The \emph{Enslaved.org} ontology about historic person and events data on the transatlantic slave trade \cite{enslaved-ont} has modules such as \emph{Event}, \emph{Place}, \emph{Provenance}, \emph{PersonRecord}, \emph{AgeRecord}, etc. Modules can overlap or even be nested, and there are, on purpose, no precise rules which ontology pieces should be thought of as being part of a specific module. Rather, it has to \emph{make sense} from several perspectives, including the domain experts' perspectives, the use case perspectives, and the perspectives of the available data relevant for the use case and ontology. As such, the modules provide \emph{conceptual bridges} between human expert conceptualization and data reality \cite{modont}.

Our approach called MOMo for \emph{Modular Ontology Modeling}, also includes a step-by-step ontology design methodology geared towards humans, a module (and pattern) description language OPLa expressed in OWL \cite{opla}, ontology design pattern libraries \cite{modl}, and a Prot\'eg\'e plug-in for ontology development \cite{comodide2,comodide-eval} -- we mention these just in passing, and details can be found in \cite{momo}; for this paper, the modular structure of the resulting ontologies themselves plus the fact that modules in OPLa annotated modular ontologies be easily be identified programmatically, are really the most important.

Knowledge graphs whose schema is given as a MOMo modular ontology then naturally inherit the modular structure from the ontology: The corresponding graph (ABox) module consists of all those ABox statements that are fully within a given ontology module. 

MOMo was designed before the advent of LLMs, geared primarily towards human ontology engineers. However it has also always been on our minds that it should also be helpful for working towards (semi-)automation of hard KGOE tasks. LLMs may now provide the opportunity for the Semantic Web community to close the remaining gaps.

%%%%%%%%%%%%%%%%%%%%%%%%%%%%%%%%%%%%%%%%%%%%%%%%%%%%%%%%%%%%%%%%%%
\section{Modular LLM-based KGOE}
\label{sec:modular}
%%%%%
We will now look, in turn, at the key KGOE tasks identified in the introduction and discuss each of them from the perspective of modularity-driven LLM-based KGOE. Specifically, we address the different tasks in essentially in order of abstraction, i.e., from knowledge to data. We note of course that inroads made in any tasks easily have the capacity to impact and improve outcomes in the others.

%%%%%%%%%%%%%%%%%%%%%%%%%%%%%%%%%%%%%%%%%%%%%%%%%%%%%%%%%%%%%%%%%%
\subsection{Ontology Modeling, Extension and Modification}
\label{ssec:ont-mem}
%%%%%
Automated ontology design -- often referred to as Ontology Learning -- has been investigated primarily from a traditional (i.e., pre-LLM) Natural Language Processing (NLP) perspective, as a possible way to address the knowledge acquisition bottleneck with much of the methods established in the early aughts; see, e.g., \cite{ol-cr,ol-sw,ol-fca,text2onto,DBLP:series/faia/2008-167,ol-handbook}. However, as the proliferation of early ODP-based methodologies (e.g., eXtreme Design \cite{extreme1}) grew, newer attempts to accomplish ontology learning emerged, leveraging identification of candidate patterns and subsequent human intervention \cite{ol-odp-diss}. 

In remaining pre-LLM years, ontology learning incrementally advanced from these early achievements and established techniques, generally through incremental evolution, inductive logic programming and both linguistic- and statistical-focused NLP \cite{ol-survey,ol-methods}.\footnote{Indeed, the capacity of modern hardware allowed some techniques to newly shine.}
However, results to date have been rather mixed, with resulting ontologies far from being able to compare with carefully crafted ontologies by domain and ontology engineering experts. The promise of LLMs now in this context comes from the fact that they are very powerful NLP tools that approximately capture a wide swath of human (expert) knowledge and can be coaxed by good prompting to even provide this knowledge in a structured form, such as expressed in a formal language over a given vocabulary.

Indeed, some attempts have already been made for LLMs, with some middling success \cite{olaf,llms4ol}. At this stage, LLMs are too limited in commonsense and reasoning power to in a fully automated fashion. Yet, we can thus expect that LLMs should enable a human ontology engineers and domain experts to first draft and then finalize suitable ontologies much more quickly than before.

It appears to us that specific aspects of the MOMo methodology, and generally the modularity idea, should further an LLM-based approach even more:
\begin{enumerate}
    \item MOMo is based on a principled use of (high-quality) ontology design patterns (ODPs). Developing ODPs is arguably easier than developing full-fledged ontologies, i.e. in a first step LLMs can be used to develop ODP libraries, which then in turn can be made use of in LLM-based ontology design. As a first step, we have used an LLM to generate hundreds of simple ``commonsense'' patterns for common concepts \cite{csmodl}. These micropatterns -- so called due to their shallow semantics and, even for ODPs, simple structure -- have been organized into a design library \cite{modl}, which allows for programmatic access, such as through a RAG \cite{lewis2020retrieval} system. Due to their simplicity, they are easily instantiated into modules \cite{template} and connected together in a modular fashion.
    \item MOMo provides a step-by-step ontology design process that breaks down the complex ontology modeling task into clearly delineated pieces, each of which should be easier to automate than going one-shot from base data (or texts) to an ontology.
\end{enumerate}
The situation for extension and modification of a MOMo ontology is rather similar -- in the MOMo approach an ontology is designed module-by-module (with possible modification of earlier developed modules while progressing), i.e. extension and modification are already part of the modeling process.

%%%%%%%%%%%%%%%%%%%%%%%%%%%%%%%%%%%%%%%%%%%%%%%%%%%%%%%%%%%%%%%%%
\subsection{Ontology Alignment}
\label{ssec:ont-align}
%%%%%
Our experience with LLM-based modular ontology alignment was already conveyed above, in Section \ref{sec:divide}. Ontology Alignment is a core task for ontology-based data integration with a long-standing corresponding research community, benchmarks and annual performance competitions.\footnote{See \url{https://om.ontologymatching.org/} for pointers.} The community has mostly focused on full automation (as opposed to human-in-the-loop semi-automation) and on so-called \emph{simple} alignment tasks, i.e. the creation of one-to-one class mappings (and, with much less emphasis, one-to-one property mappings). The need for \emph{complex} ontology alignment -- i.e., the creation of mapping \emph{rules} that go beyond one-to-one mapping -- has long been recognized, but it has only come into more focus in the community about 10 years ago, with very limited results (as pointed out in Section~\ref{sec:divide}, see also \cite{DBLP:conf/semweb/PourAAFFHHJJKKL20,DBLP:conf/semweb/PourAAAFFFHHHHI21}. Regretfully (we think), since the advent of LLMs focus has mostly shifted back to simple alignments, just now with the support of LLMs. Indeed, LLM-based \emph{simple} ontology alignment has already been investigated with good results \cite{DBLP:conf/om2/NorouziMH23,DBLP:conf/kcap/HertlingP23}. 

We would be negligent to not point out that we also observe issues with some of the established current benchmarks and methods. For example, in \cite{conference-bm} a simple alignment benchmark was re-evaluated, finding that even humans disagree on a significant part of the benchmark, essentially demonstrating that the benchmark ontologies simply do not carry enough information such that even humans can solve it perfectly. In \cite{cheatham13}, it was demonstrated that most of the functionality of simple alignment systems can be obtained by using only string similarity metrics. At the same time, when submitting papers for publication that described complex alignment benchmarks that came out of real data integration work \cite{geolink-bm,enslaved-bm}, some reviewers dismissively pointed out that they would not be good as benchmarks because they would be far beyond anything that could currently be done.\footnote{One of the reasons for creating these benchmarks was precisely to point out that the state of the art is way behind practical needs.}

For some years, progress on the complex alignment task stalled \cite{oaei-2020}, even as new benchmarks were released \cite{complex-eval}. However, some new techniques, even incorporating LLMs have shown some promise. For example, the composition of (geometric) language embeddings cast from a KG or sourced from an LLM, can be shown to have correspondences in the latent space \cite{DBLP:conf/ecai/SilvaFP24}. Yet, the performance is still relatively low, ranging from 0.49--0.69 for Semantic F1-score, depending on the ontologies. As described above in Section \ref{sec:divide}, it was demonstrated in \cite{KGSWC-OA-2024} that modularity provides significant performance improvements to LLM-based complex alignment, in this case with 104 out of 109 (i.e., 95\%) target alignment mappings correctly identified on the GeoLink benchmark when taking modularity into account.

%%%%%%%%%%%%%%%%%%%%%%%%%%%%%%%%%%%%%%%%%%%%%%%%%%%%%%%%%%%%%%%%%
\subsection{Ontology Population}
\label{ssec:ont-pop}
%%%%%
In some sense, ontology population and entity co-reference resolution/disambiguation are not substantially different. On one hand there is a \emph{theoretical or unknown} entity which satisfies, or otherwise conforms to, a portion of an ontology (i.e., it should populate the ontology). On the other hand, there is a known entity described, or otherwise named, in the natural language corpus. Now the system needs to determine if those two entities \emph{match} to move forward in populating the ontology.

In this case, we can leverage two aspects of modularity for ontology population with LLMs: conceptual consistency and tight scope. That is, the conceptual consistency of a module -- based on our definition, of course -- means that the constituent classes and properties of a module somehow belong together, especially from a human perspective. This in turn provides a tighter cluster of ``sentiment'' (for lack of a better term), priming an LLM in the prompt.

On the other hand, we have seen that LLMs tend to be better at following patterns, rather than instructions, and it correlates as well to the length of the prompt \cite{levy2024tasktokensimpactinput}. Thus, we posited that attempting ontology population on a per-module basis would succeed where other attempts had poorer results.

Indeed, in a recent set of experiments, we obtained rather excellent results \cite{ont-pop-llms}. Using prompts constructed with a simple schematic representations of modules and a corresponding extraction example, an LLM was able to achieve \~90\% extraction of related triples from text, as compared to ground truth. While additional experiments are of course required, especially examining appropriate modeling characteristics for modules, we take this as an excellent indicator that modularity is of significant added value to the process.

%%%%%%%%%%%%%%%%%%%%%%%%%%%%%%%%%%%%%%%%%%%%%%%%%%%%%%%%%%%%%%%%%
\subsection{Entity Disambiguation}
\label{ssec:ent-dis}
%%%%%
As above, we note that ontology population and entity disambiguation are tightly intertwined. While ontology population focuses on the extraction of an entity and (ostensibly) comparing it to a theoretical candidate, entity disambiguation compares two known, or otherwise named, entities and must produce a value corresponding to the degree to which they are the same entity.

However, we also note that it is difficult to assess the current capabilities of LLMs to produce correct entity disambiguation due to the possibility for data leakage (i.e., the appearance of existing and gold-standard benchmarks in the testing data). For example, some benchmarks (after prompt-tuning) achieve upwards of $F_1=91$.

As successful co-reference resolution correlates with available context, per improved ability to resolve matching context (i.e., complex ontology alignment) we see immediate benefit in this parallel task of entity disambiguation. We thus posit that the added context from a tight, conceptual description, i.e., a module, will significantly improve outcomes for this KGOE task.

%%%%%%%%%%%%%%%%%%%%%%%%%%%%%%%%%%%%%%%%%%%%%%%%%%%%%%%%%%%%%%%%%
\section{Concrete Research Challenges}
\label{sec:challenges}
%%%%%
Throughout this paper so far, we have largely discussed a specific type of module, which we have been calling a \emph{conceptual module} and distinct from other uses in the literature for both ontologies \cite{modularity-overview,dl-modules} and KGs \cite{mod-kgs}. We are not aware of any work on the capabilities of LLMs for integrating for these other definitions of module. We suspect that due to their (in general) rigorous approach to the partitioning of knowledge in a logically consistent way, we may see similar benefits as when using our notion of modularity. That is to say, by somehow limiting the scope, we achieve a more human-like approach -- and one more capable of being expressed succinctly in language, and thus more appropriate for LLM-based assistance in the task. 

Of course, it is important to not altogether \emph{neglect} further research on non-modular, LLM-based KGOE; we need to understand its limitations, as well as its capabilities. Indeed, additional focus on modular or pattern-based techniques for KGOE can inform these other lines of research.

From a broader perspective, the added value of modularity that seems to become apparent for LLM-based KGOE prompts the question what other improvements to established ontology and knowledge graph paradigms would be similarly easy to make and could also provide major benefits. It is the task of the Semantic Web, Knowledge Graphs, and Ontologies research communities to look into this very question.

%%%%%%%%%%%%%%%%%%%%%%%%%%%%%%%%%%%%%%%%%%%%%%%%%%%%%%%%%%%%%%%%%
\section{Conclusion}
\label{sec:conc}
%%%%%
Knowledge Graphs and Ontologies have seen new resurgence in the era of LLMs, for example in structuring data for RAG systems \cite{kgs-llms-roadmap} or providing guide-rails to limit or prevent confabulation in textual output \cite{DBLP:conf/dsaa/GilpinPK21}. Yet, for all their established and growing importance, many of the tasks pertaining to their development, maintenance, extension, and population (to name a few) are still very difficult. It seems somehow appropriate that we incorporate LLMs into improving KGOE tasks. LLMs have shown some initial success in \emph{simple} tasks, such as the generation of simple schemas or topic extraction. 

To tackle the \emph{hard} tasks, we have proposed the use of modules -- conceptual, human-centric partitions of an ontology or schema -- that provide internal structure. These conceptual boundaries assist LLMs in a variety of tasks, seeing already success in ontology construction, (complex) alignment, and population. It seems obvious now that modularity is a \emph{missing link} for bridging human conceptualization and machine interoperability. We thus fully believe that modularity must be incorporated from the start, both structurally and in documentation, so as to further enable the various improvements to KGOE tasks we have outlined above.

%%%%%%%% Acks
\medskip

\noindent\emph{Acknowledgment.}   
This work was supported by the National Science Foundation under award 2333532 "Proto-OKN Theme 3: An Education Gateway for the Proto-OKN" (EduGate).

%%%%%
\bibliographystyle{abbrv}
\bibliography{refs}

\end{document}